\documentclass{article}

\usepackage{booktabs} % For formal tables

\usepackage{algpseudocode}
\usepackage{algorithm}
\usepackage[bookmarks=false]{hyperref}
\usepackage{graphicx}
\usepackage{multicol}
\usepackage{color}
\usepackage{authblk}

\title{Saccadic Predictive Vision Model with a Fovea}
\author[1,2]{Michael Hazoglou}
\author[1,2]{Todd Hylton}
\affil[1]{Contextual Robotics Institute, University of California San Diego}
\affil[2]{Department of Electrical and Computer Engineering, University of California San Diego}

\begin{document}
\maketitle
   
\begin{abstract}
	We propose a model that emulates saccades, the rapid movements of the eye, called the Error Saccade Model, based on the prediction error of the Predictive Vision Model (PVM). The Error Saccade Model carries out movements of the model's field of view to regions with the highest prediction error. Comparisons of the Error Saccade Model on Predictive Vision Models with and without a fovea show that a fovea-like structure in the input level of the PVM improves the Error Saccade Model's ability to pursue detailed objects in its view. We hypothesize that the improvement is due to poorer resolution in the periphery causing higher prediction error when an object passes, triggering a saccade to the next location. 
\end{abstract}

\section{Introduction}
\label{intro}
Machine learning techniques have exceeded human ability in certain specialized areas such as games \cite{2013arXiv1312.5602M,silver2016mastering,campbell2002deep,lewis2012game} and classification problems but not in domains like perception and mobility in real world environments, which require the ability to generalize broadly. In the case of games, where the full state of the system is given, there no need for generalization as the algorithm can specify the optimal action for the every state. Classification problems in computer vision are limited by their lack of generalization beyond the set of labels and data used to train them. Examples of this would be labels such as cars, airplanes, building, boats, etc. with pictures of these objects from the exterior, a deep neural network would struggle to correctly label these objects if they were in components or from their interiors. The ability to move around a car gives more information than what can be extracted from a single photo due to spatiotemporal information and context, allowing a person to make generalizations about cars. We propose a biologically inspired visual model that carries out the simplest type of mobility - saccades - coupled to a visual model that learns by prediction on real world video.

Saccades are rapid eye movements made between two fixation points. As you read this paper your eyes will fixate on one word, and you will not be able read beyond a few words into your peripheral vision. Human eyes do not simultaneously process large fields of view, but selectively scan the environment to form a complete scene. The human visual system has features that allow it to exceed it's own physical limitations such as it's spacial Nyquist limit \cite{EdgeDetectionWithoutOverlap} by using a combination of specialized eye movements and effective processing of the sensory input. It is known that without eye movements vision fades \cite{Coppola8001,martinez2004role}, which can even be seen outside of the lab in Troxler's fading. We implement saccades by integrating a visual system model called the Predictive Vision Model (PVM) \cite{DBLP:journals/corr/PiekniewskiLPRF16} with a new model of saccades using the prediction error of the PVM and a fovea like structure in the input level of the model.  Because of their limited field-of view, our models receive only a portion of the video input to which they are exposed at any given time and must saccade to see other portions of the input. By comparing models with and without a foveal region, we show that model the with a fovea is better at spontaneously following detailed objects that come into view.

\section{Predictive Vision Model with a Fovea}
\label{PVMwithFovea}
\subsection{Predictive Vision Model}
		
\begin{figure}
		\centering
		\includegraphics[width=\columnwidth]{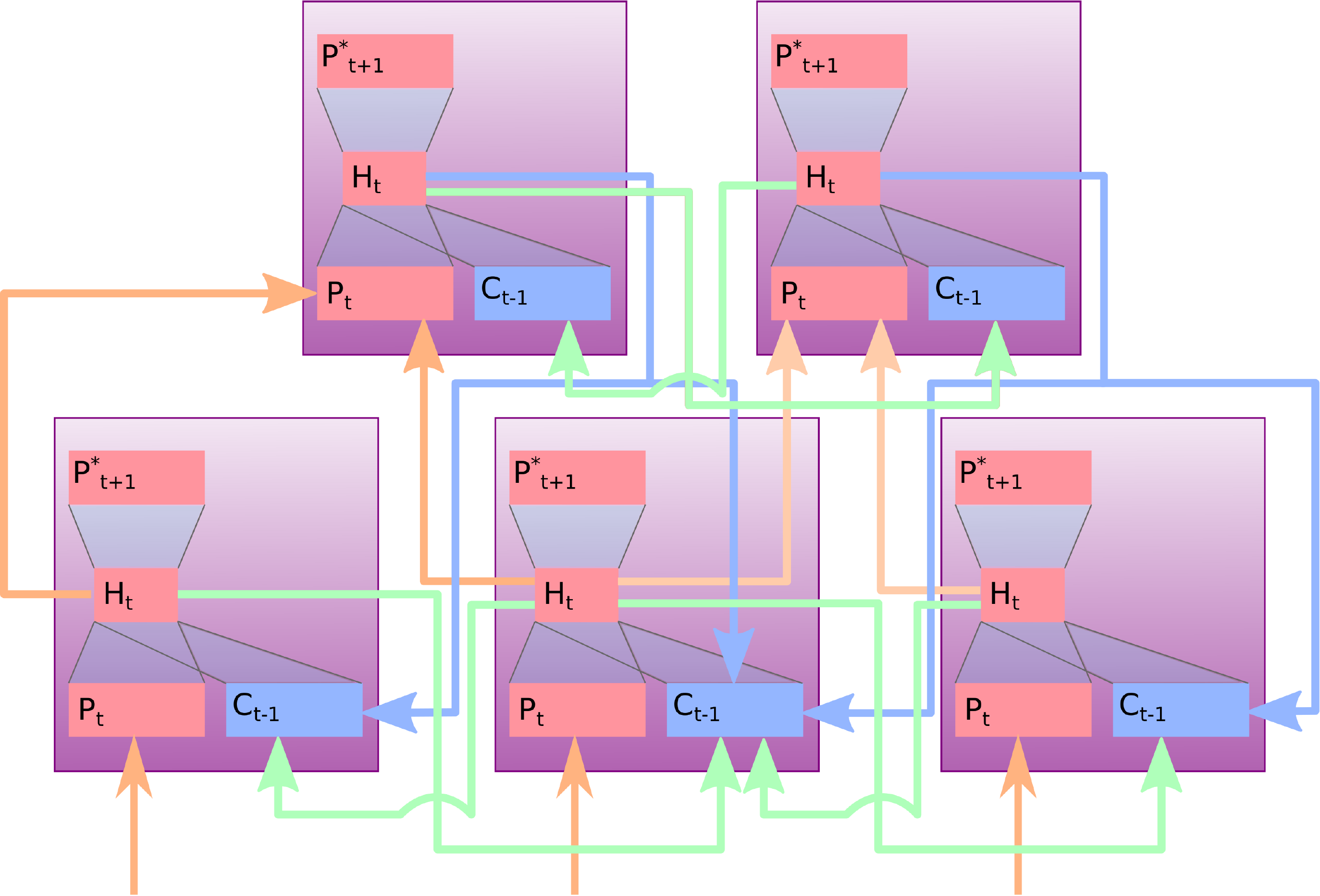}
		\caption{An example of the PVM hierarchy used in \protect\cite{DBLP:journals/corr/PiekniewskiLPRF16}. The purple boxes represent the PVM units. Orange arrows represent the inputs into the units. The blue arrows show the flow of context from superior units while the green arrows show the flow of lateral context. The primary signal (bottom large red box) and context (in blue) is compressed to generate the (hidden) state of the unit which is show as the smaller red box from which a prediction (top large red box) is made for the next input the unit will receive. Since the inputs to the units are fed one at a time, time derivatives, error in the previous prediction and time averages (integrals) are also factored in the hidden state calculation.}
		\label{BasePVMhierarchy}
	\end{figure}

The Predictive Vision Model (PVM) is a hierarchical, recursive network of predictive learning units and has been successfully used as a tracker that compares well with state of the art, bespoke tracking models \cite{DBLP:journals/corr/PiekniewskiLPRF16}. The units used in \cite{DBLP:journals/corr/PiekniewskiLPRF16} and in the models described herein are three-layer perceptrons using sigmoid activation neurons, though other unit types such as CNN or LSTMs based units might also be used.  Each unit learns to predict future inputs and uses prediction error as a training signal. For example, at the lowest level of the hierarchy, raw video input to the input (first) layer of the unit's three-layer perceptron at frame time $t$ might be used to predict the next frame that the input level will receive at time $t+1$.  Prediction errors computed in the top (third) perceptron layer are used to train the hidden (second) perceptron layer, using a backpropagation algorithm. Hidden layer activations become the next time step inputs for higher level units of the hierarchy as well as "contextual" inputs to units in the same or lower levels.  Additional inputs to units may include integrals, derivatives, hidden layer activations, and additional earlier time inputs and context. The details of these calculations are summarized in Table~\ref{PVMunitCalculation}.  The input video is broken into tiles that align with the PVM units in first level of the hierarchy. The first level feeds its hidden states to the second level as an input signal for the units in that level, and the second level feeds its hidden state to the third level, etc. PVM units in the higher levels are referred to as superior units. The basic hierarchy of the PVM is shown in Fig.~\ref{BasePVMhierarchy}. The predictive learning technique of the PVM means that its training can also be fully unsupervised - it learns using only the sequence of frames presented on its inputs. The PVM can also be embarrassingly parallelized as individual PVM units run independently using context from the previous time step. A scheme is shown in algorithm~\ref{PVMunit} for how the PVM hierarchy can be run in parallel.

	\begin{algorithm}
		\caption{Predictive Vision Model Unit (runs in parallel for all units)}
		\label{PVMunit}
		\begin{algorithmic}
			\While{$True$}
			\State $frame\_gpu\_arr \gets \mbox{To\_Gpu}(frame)$
			\State $signal \gets \mbox{Map\_Frame}(frame\_gpu\_arr)$
			\State $precomputed\_features \gets \mbox{Precompute\_Features}(signal)$ 
			\State \Comment{Contains the derivative, integral and prediction error}
			\State Synchronize(streams)
			\State $input \gets \newline Concatenate(signal, precomputed\_features, context)$
			\State $signal\_prediction \gets Make\_Prediction(input)$
			\State $p\_error \gets \newline \mbox{Calculate\_Prediction\_Error}(signal\_prediction, signal)$
			\State $\mbox{Train}(p\_error)$ \Comment{Backpropagation is removed when training is complete}
			\State Synchronize(streams)
			\State $context \gets \mbox{collect\_current\_context}()$
			\State Synchronize(streams)
			\EndWhile
		\end{algorithmic}
	\end{algorithm}
	
	\begin{table}
		\begin{tabular}{| p{4cm} | c | p{5.2cm} |}
			\hline 
			Layer & Symbol & Definition \\ \hline
			\multicolumn{3}{|c|}{\textbf{Inputs}} \\ \hline
			Signal & $P_t$ & Fan-in from inferior level or raw video tile \\ \hline
			Integral & $I_t$ & $\tau I_{t-1} + (1-\tau)P_t$\\ \hline
			Derivative & $D_{t/t-1}$ & $0.5 + (P_t - P_{t-1})/2$ \\ \hline
			Previous Prediction Error & $E_t$ & $0.5 + (P^*_t - P_t)/2$ \\ \hline
			Context & $C_{t-1}$ & concat[$H_t \dots$] from Hidden of self/lateral/superior/topmost units \\ \hline
			\multicolumn{3}{| c |}{\textbf{Output}} \\ \hline
			Hidden & $H_t$ & $\sigma(W_h \cdot [P_t; D_{t/t-1};I_t;E_{t};C_t] + b_h) $ \\ \hline
			\multicolumn{3}{| c |}{\textbf{Predictions}} \\ \hline
			Predicted Signal & $P^*_{t+1}$ & $\sigma(W_p \cdot H_t + b_p) $ \\ \hline 
		\end{tabular}
		\caption{Summary of the PVM unit. Each unit consists of a three layer MLP with sigmoid action neurons. The indices represent the time step for the respective value.}
		\label{PVMunitCalculation}
	\end{table}
	
\subsection{Structure of the Foveal Region}
\label{FoveaStructure}
\begin{figure}
		\centering
		\includegraphics[width=.7\columnwidth]{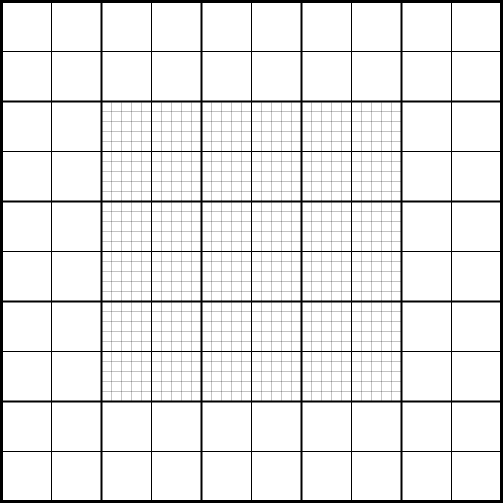}
		\caption{An example of how the input is divided amongst the PVM units in the input level of the model to create a fovea like structure. Each square here represent the tile that a PVM unit is getting as an input from the viewed region of the frame. The lateral connectivity between units is such that if the borders or their tiles touch they are connected.}
		\label{InputlevelStructure}
	\end{figure}

\begin{figure}
		\centering
		\includegraphics[width=.8\columnwidth]{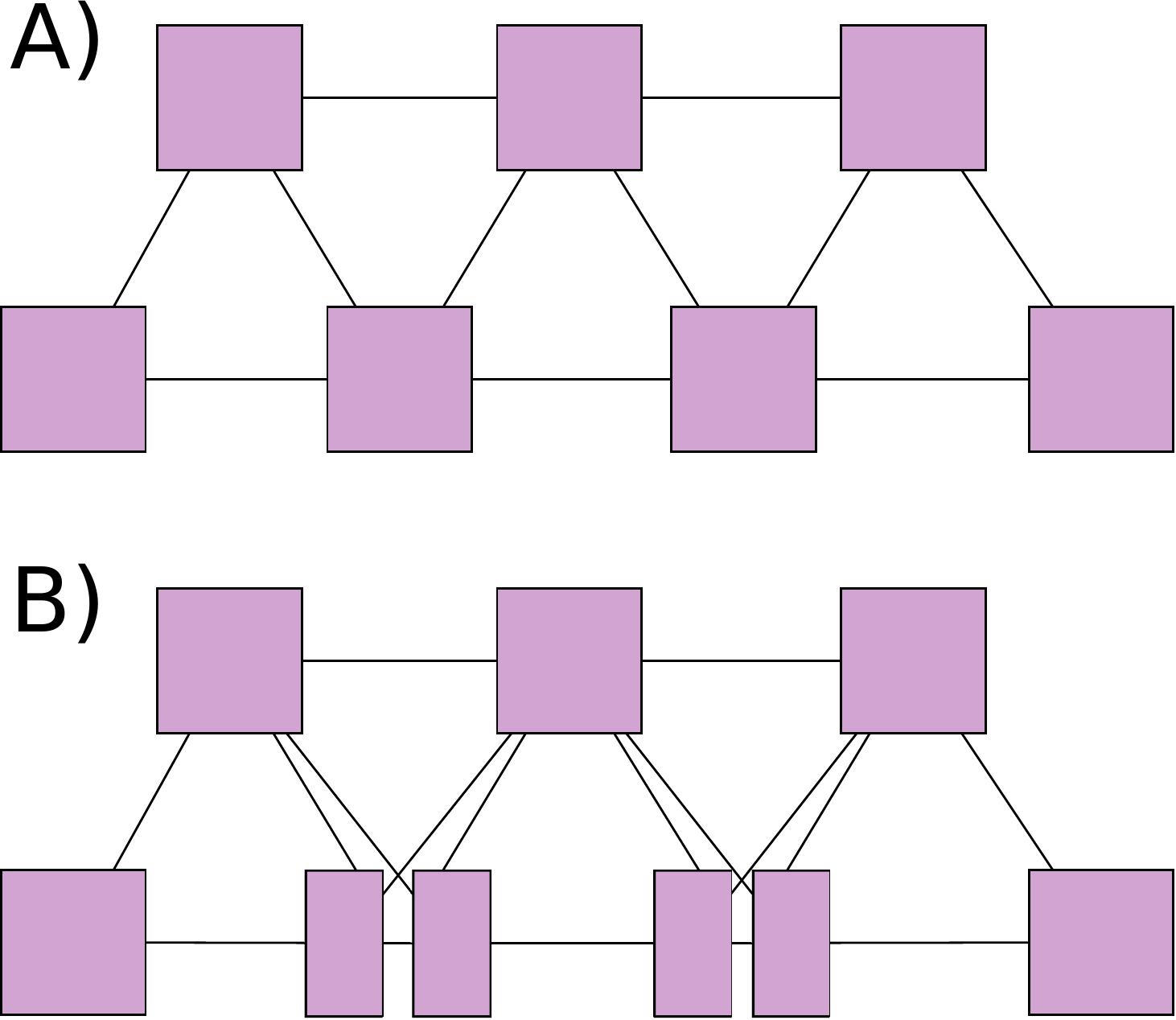}
		\caption{ A) An example of a slice of the PVM hierarchy each square here is a unit the lines between units represent which units share their hidden state (i.e. the units are connected). B) An example of the previous slice's connectivity after subdividing the middle two units. The lateral connections in the first level occurs when the boundaries of units inputs are touching (see Fig.~\ref{InputlevelStructure}). The superior connections between the first and second level occur if the unit that was subdivided was initially connected to a unit in the adjacent level.}
		\label{ConnectivityBetweenlevels}
	\end{figure}

	In the work that follows we will compare the performance of foveated and non-foveated PVM hierarchies.  In the non-foveated hierarchy, the input level is uniformly tiled with identical PVM units, each unit receiving the same number of input pixels.  In the foveated models, however, the hierarchy is changed such that the input level has a higher density of PVM units in the central region. We implemented this idea by subdividing the central region of the originally uniform rectangular grid of the non-foveated input level into a smaller grid. An example of such a structure is drawn in Fig.~\ref{InputlevelStructure}. The connectivity of the PVM units between and within levels in the foveated hierarchy is the same as in the non-foveated hierarchy with the exception that lateral connections between the units in the input level are connected if the boundaries of the tiles of the input fed to the PVM units are adjacent. The superior connections of units in the input (first) level and units in the adjacent (second) level occur if the unit in the input level originated from a subdivision that was connected to a unit in the adjacent level (see Fig. \ref{ConnectivityBetweenlevels}).

\section{Saccades are made to regions of high prediction error: The Error Saccade Model}
\label{SaccadeErrorModel}
	It has been suggested that fixational eye movements are key to edge detection as well as acuity beyond the spacial Nyquist limit see \cite{EdgeDetectionWithoutOverlap} and references therein. For the PVM the prediction error of the model is a key quantity in edge detection. The boundaries of moving objects would result in high prediction errors occurring and work as a type of edge detection without the need for overlapping PVM units or convolution. In the case of the PVM getting a constant visual stimulus the prediction error will tend towards zero for all values, the model will not detect edges. This is analogous to vision fading which occurs when fixational eye movements are mitigated so that a visual stimulus remains constant on the retina \cite{Coppola8001,martinez2004role,Sharpe1972} where the boundaries between different objects vanish and blur into single color.

	Using the prediction error we can calculate the total square error in rectangular sliding windows of the field of the model's vision. The field of view will move to the current region with the highest total error if it exceeds the time average of the maximum total errors of previous windows. The movement of the field of view is given by the same dynamics of a damped harmonic oscillator. The model then updates the time average of the maximum error with the current maximum. The detailed algorithm is shown in Algorithm \ref{ErrorSaccade}.

	The choice of a damped, isotropic, harmonic motion model for saccades was chosen for its simplicity and motivated by the facts that any system with a stable equilibrium can be approximated by a harmonic potential and dissipation plays a role in any real physical system. In some ways this model is consistent with some experimental observations of the movements of the iris and lens; they are also known to follow the motion of a damped harmonic oscillator, but they are not isotropic (more horizontal movement than vertical is observed) \cite{10.1371/journal.pone.0095764}. The model also includes a small noise term added to the position of the field of view to create a constantly changing visual stimulus that avoids prediction error fading and makes edges more apparent. This small noise term is made to approximate fixational eye movements \cite{martinez2004role,EdgeDetectionWithoutOverlap}.

	\begin{algorithm}
		\caption{Error Saccade Model}
		\label{ErrorSaccade}
		\begin{algorithmic}
			\State $\omega \Delta t \gets 0.8$ \Comment{Related to the strength of the forcing to the fixation point}
			\State $\gamma \gets 0.9$ \Comment{Any underdamped value close to 1 should work well}
			\State $threshold \gets 0$
			\For{$frame$ in $Frames$}
            \State $subframe \gets frame[y_t:y_t + h, x_t: x_t + w, :]$ \Comment{Cutting out the part of the frame in the model's view}
        	\State $Prediction, \, Error, \, Hiddens \gets PVM\_forward(subframe)$
            \State $SquareError \gets Error * Error$
			\State $PatchTotalSquareError \gets \mbox{PatchWiseSum}(SquareError)$
			\State $MaxErrorValue \gets \mbox{max}(PatchTotalSquareError)$
			\State $MaxErrorIdx \gets \mbox{argmax}(PatchTotalSquareError)$
			\If {$MaxErrorValue > threshold$}
			\State $x_{fix} \gets MaxErrorIdx.x$
			\State $y_{fix} \gets MaxErrorIdx.y$
			\EndIf
			\State $x_{t+1} \gets \left\lfloor\frac{[2 - (\omega \Delta t)^2] x_t + (\gamma \omega \Delta t - 1) x_{t-1} + \omega^2 \Delta t^2 x_{fix}}{1 + \gamma \omega \Delta t} \right\rceil + randint(-l, l)$
			\State $y_{t+1} \gets \left\lfloor\frac{[2 - (\omega \Delta t)^2] y_t + (\gamma \omega \Delta t - 1) y_{t-1} + \omega^2 \Delta t^2 y_{fix}}{1 + \gamma \omega \Delta t} \right\rceil + randint(-l, l)$
			\State $x_{t-1} \gets x_{t}$
			\State $x_{t} \gets x_{t+1}$
			\State $y_{t-1} \gets y_{t}$
			\State $y_{t} \gets y_{t+1}$
			\State $threshold \gets (1 - \tau) \times threshold + \tau \times MaxErrorValue$ \Comment{$0 \leq \tau \leq 1$}
			\EndFor
		\end{algorithmic}
	\end{algorithm}
	
	\begin{figure}[h]
		\includegraphics[width=\columnwidth]{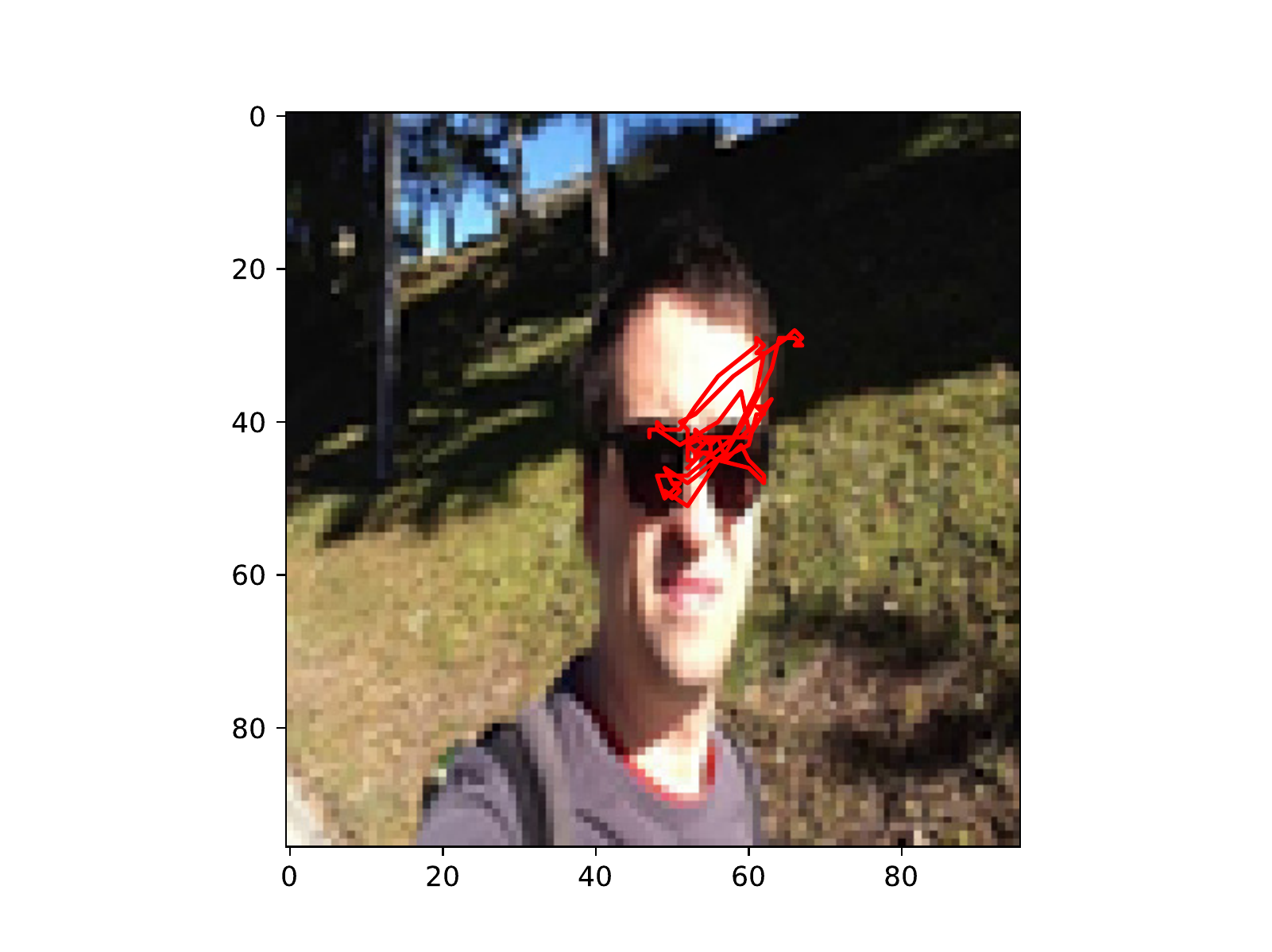}
		\caption{A trace of a frame from the testing set in \protect\cite{DBLP:journals/corr/PiekniewskiLPRF16} using a foveated PVM that can only view a 32 by 32 pixel square of the image at any given time. The red line represents the trajectory of the center of the model's field of view over 100 iterations.}
		\label{FaceTrace}
	\end{figure}
	
	 In line with tradition in this field  \cite{yarbus1967eye,friston2012perceptions,najemnik2005optimal,DBLP:journals/corr/MnihHGK14}  we have produced a trace of a human face using a static image as an input. A PVM model accepting a 32 by 32 pixel portion of the image as the input, with a 6 level hierarchy of PVM units in a square grid arrangement with edge length 16, 8, 4, 3, 2, and 1 where used. It was trained using the video training set from the original paper proposing the PVM \cite{DBLP:journals/corr/PiekniewskiLPRF16}. The results of the trace can be seen in Fig.~\ref{FaceTrace}. The error saccade model spends most of its time tracing left and right across the man's face alternating between his left and right eye.  We note that this is done without the bias of many years of evolutionary pressure to recognize faces or any other feature bias built into the model. This model seems to work as a plausible first approximation to saccades. Video of the same model in action on the testing set from the original PVM paper can be seen here \url{https://youtu.be/-W3-DSsNWHA}.

	\section{Fovea-like structures in the input level improves the ability of saccades to follow complex objects}
	\label{FoveaImprovesSmoothPursuit}
	We have carried out experiments with three different instances of the PVM in order to investigate the effect of adding foveal structure to the models. All instances were trained in an unsupervised manner for 3 million frames at a learning rate of 0.01 on the training set used in \cite{DBLP:journals/corr/PiekniewskiLPRF16}. Each model is a 6 level hierarchy with identical structure in levels 2-6 but with differing structure in the input level (level 1).  The upper levels in each version consist of 8 by 8, 4 by 4, 3 by 3, 2 by 2, and 1 units successively. Each PVM unit has a hidden layer of 8 neurons.  The input levels are as follows
	
	\begin{itemize}
		\item Base model: The first level is 16 by 16 grid of PVM units taking 2 pixel by 2 pixels chunks of the input frame. 
		\item Foveated model: As compared to the base model, the input is altered by taking an 8 by 8 grid of units in the center of the base model's first level and breaking them into 4 units (2 by 2) that each take a single pixel of the input image.  The connectivity to the second level as described in section~\ref{FoveaStructure}.  The central high density region is the "fovea".
		\item Uniform High Resolution (UHR) model:   As compared to the base model, this model subdivides all units into 4 (2 by 2) units that each take a single pixel of the input. This makes the entire field of view the same density as the fovea of the foveated model.
	\end{itemize}
	
	These models were then tested on a set of 176 by 99 pixel videos where a small car moves around on a carpet. Each of the models can only view a 32 by 32 pixel region of the frame, with the error saccade model using a sliding window of 3 by 3 pixels to compute prediction error in the window. This video was chosen for the contrast it provides between the complex features of the car and its operator versus the relatively uniform features of the floor and walls surrounding them. In order to quantify the ability of the models to saccade toward complex features, the image entropy of each image frame of the testing video was calculated for each color channel using Python's skimage module with a disk morphology of radius 5 pixels. Entropies of the independent color channels where added together for the total image entropy. Image entropy $S$ is defined by
	
	\begin{equation}
		S = -\sum_{i, c} p_{i, c} \log_2 p_{i,c}
		\label{Entropy_equation}
	\end{equation}
	where $p_{i,c}$ is the normalized distribution of color $c$ with integer value $i$ within a certain patch. From the above equation one can see that image entropy would be zero for a uniform image (all a single color) and is larger the more distributed the color values are in a patch of the image. The image entropy was used as a metric of local image complexity to evaluate the ability of the different models to saccade toward complex features and objects.
	
\begin{figure}
		\includegraphics[width=\columnwidth]{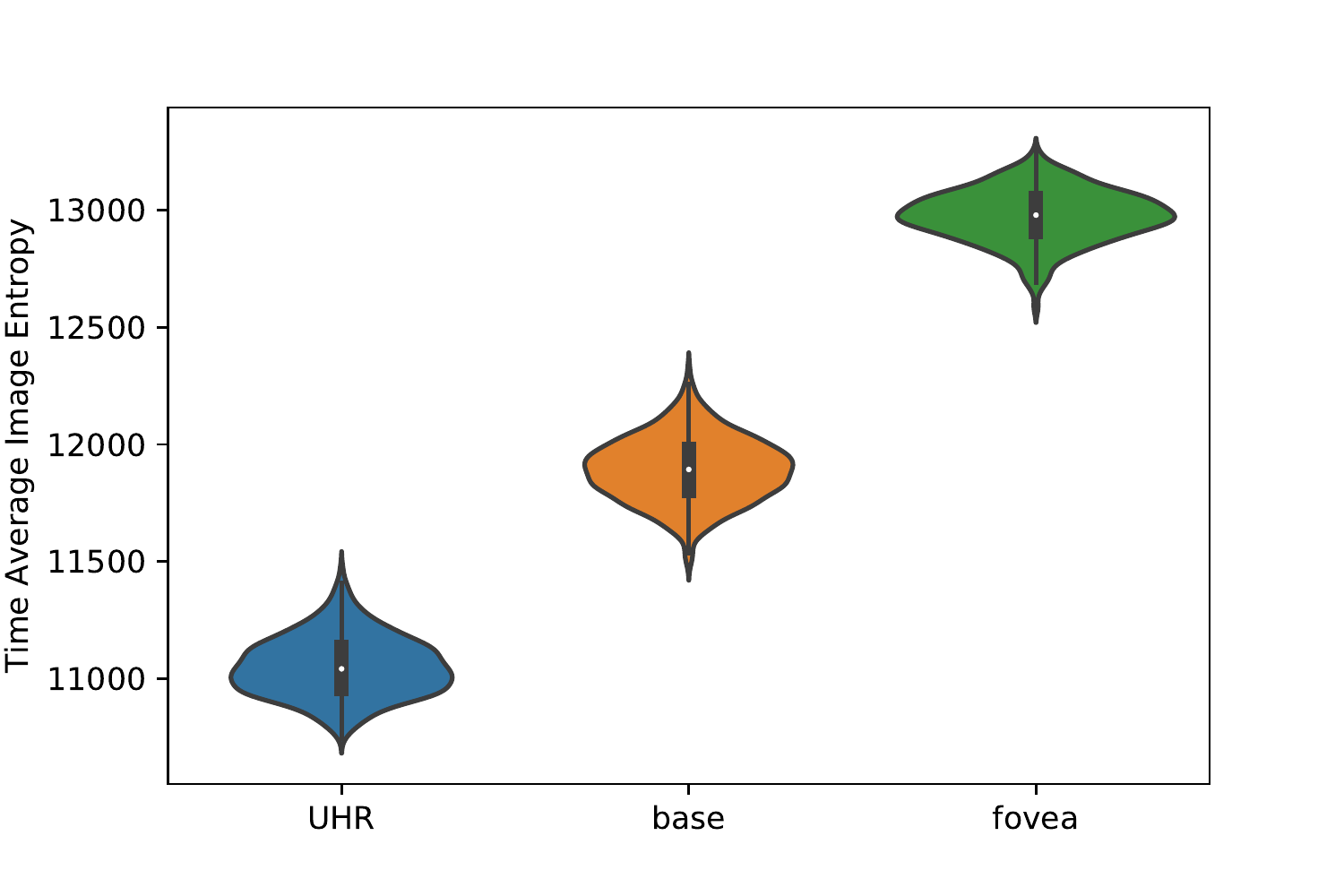}
		\caption{This is a violin plot comparing the distribution of image entropy averaged across all frames in the testing set. The distributions were estimated using 1000 trials.}
		\label{ViolinPlotOfImageEntropy}
	\end{figure}

	We ran 1000 trials across the testing set and compared the average image entropy across frames (time-average) that each of models had in its field of view. The results are summarized in a violin plot in Fig.~\ref{ViolinPlotOfImageEntropy}.The results clearly show that the foveated model spends a significantly higher amount of time with higher entropy portions of the video in its view, and that the base and UHR models have rather significant overlap in terms of their performance.
	
\begin{figure*}[h]
			\centering
			\href{https://www.youtube.com/watch?v=Fp5Wv80O4Qo}{
				\includegraphics[width=\textwidth]{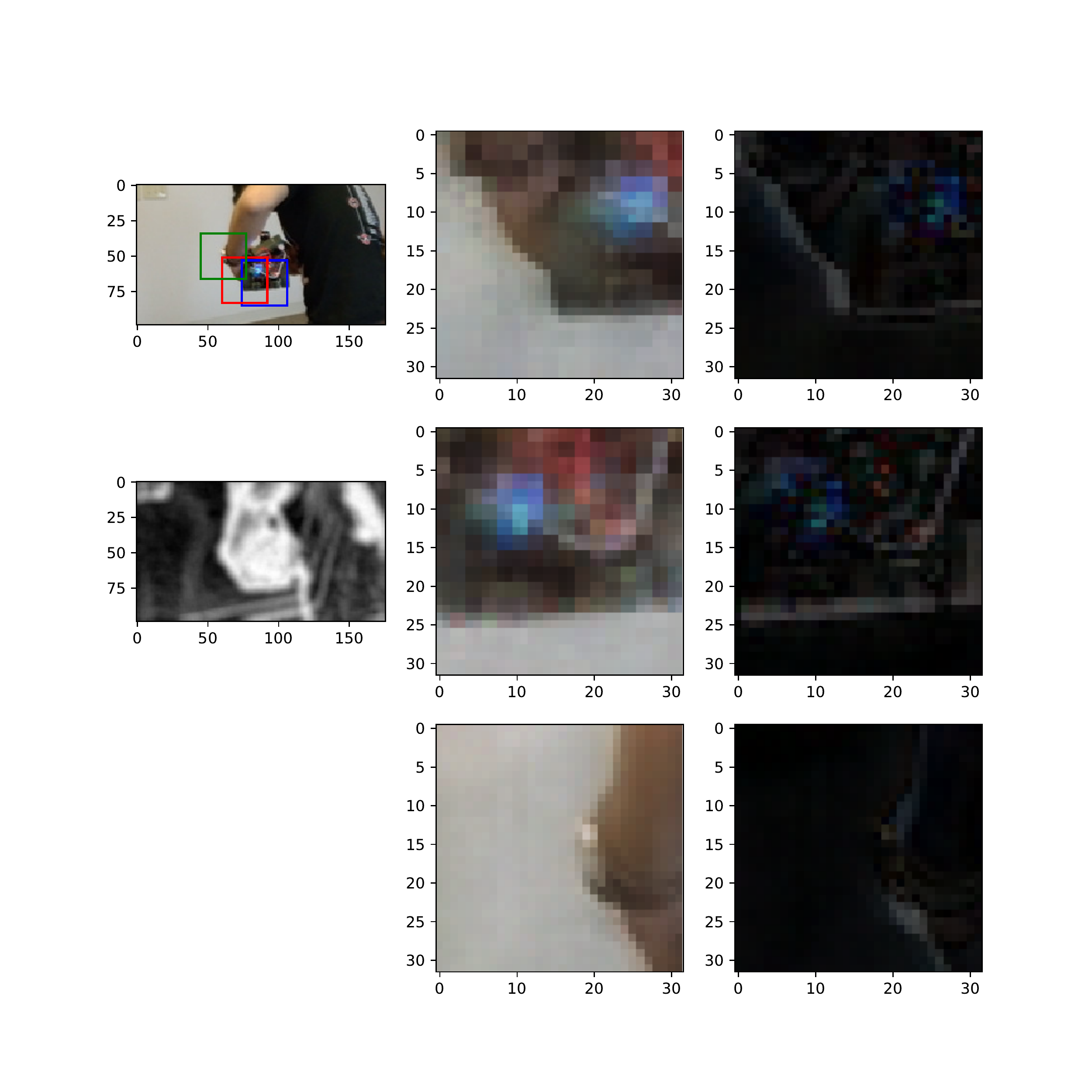}
			}
			\caption{The left column is frame from the video (top) and the image entropy (middle), the red, blue and green boxes are the foveated, base and UHR models, respectively. Center column contains the predictions, and the right column contains the prediction errors of the foveated (top), base (middle) and UHR (bottom) models. The figure links to a video.}
			\label{Video}
		\end{figure*}

	When looking at video of a trial (see Fig.~\ref{Video}) it is apparent that the foveated model is more easily attracted to complex objects like the car when they cross the periphery of its field of view than and the base and UHR models. We think this is due to momentarily higher prediction error in the periphery combined with the higher resolution in the fovea resulting in a saccade toward the object followed by its smooth pursuit.
	
    \newpage
	\begin{multicols}{2}
	\end{multicols}
	
	\section{Conclusion}
	Here we have demonstrated a closed-loop, unsupervised, saccading vision system that reduces prediction errors of a predictive vision model by moving its field of view to regions of high error.  We have also shown that imposing a fovea-like region in the model can improve its ability to smoothly pursue detailed objects. Future work will involve testing this model out on a camera with a pan tilt unit to allow for movement and to combine the error saccade model with a model of object permanence and an the Bayesian search method proposed by \cite{najemnik2005optimal}.
	
	\section*{Acknowledgements}
	MH and TH would like to thank Filip Piekniewski for useful discussion. MH would like to thank Teofilo Erin Zosa IV for allowing taping of his robot. This research was supported by Department of Energy ASCR Award DE-SC0017027.

	\bibliographystyle{natbib}
	\bibliography{ICONS2018.bib}

\end{document}